\documentclass[twoside,leqno,twocolumn]{article}

\usepackage[letterpaper]{geometry}

\usepackage{ltexpprt}
\usepackage{hyperref}

\usepackage{url}            
\usepackage{booktabs}       
\usepackage{amsfonts}       
\usepackage{nicefrac}       
\usepackage{microtype}      
\usepackage{xcolor}

\usepackage{graphicx} 
\usepackage{subcaption}
\usepackage{wrapfig}
\usepackage{multirow}
\usepackage[normalem]{ulem}
\useunder{\uline}{\ul}{}
\usepackage{makecell}
\usepackage{amsmath}

\usepackage{comment}

\begin{document}

\title{\Large Automated Learning of Semantic Embedding Representations for Diffusion Models}

\author{Limai Jiang\thanks{Shenzhen Institutes of Advanced Technology, Chinese Academy of Sciences.} \thanks{University of Chinese Academy of Sciences.} \and 
Yunpeng Cai\textsuperscript{$\ast$}\thanks{Corresponding author. Email: \{lm.jiang2, yp.cai\}@siat.ac.cn}\\
}

\date{}

\maketitle

\fancyfoot[R]{\scriptsize{Copyright \textcopyright\ 2025 by SIAM\\
Unauthorized reproduction of this article is prohibited}}

\providecommand{\keywords}[1]
{
	\small	
	\noindent\textbf{Keywords:} #1
}

\begin{abstract} \small\baselineskip=9pt 	
Generative models capture the true distribution of data, yielding semantically rich representations. Denoising diffusion models (DDMs) exhibit superior generative capabilities, though efficient representation learning for them are lacking. In this work, we employ a multi-level denoising autoencoder framework to expand the representation capacity of DDMs, which introduces sequentially consistent Diffusion Transformers and an additional timestep-dependent encoder to acquire embedding representations on the denoising Markov chain through self-conditional diffusion learning. Intuitively, the encoder, conditioned on the entire diffusion process, compresses high-dimensional data into directional vectors in latent under different noise levels, facilitating the learning of image embeddings across all timesteps. To verify the semantic adequacy of embeddings generated through this approach, extensive experiments are conducted on various datasets, demonstrating that optimally learned embeddings by DDMs surpass state-of-the-art self-supervised representation learning methods in most cases, achieving remarkable discriminative semantic representation quality. Our work justifies that DDMs are not only suitable for generative tasks, but also potentially advantageous for general-purpose deep learning applications.

\end{abstract}

\section{Introduction}

Denoising Diffusion Models (DDMs)~\cite{1,2} represent a recent class of likelihood-based models that model the distribution of data by estimating noise and iteratively denoising it~\cite{3}. They have demonstrated extremely superior performance in generative tasks such as image editing~\cite{4}, inpainting~\cite{7,8}, style transfer~\cite{9,10}, video editing~\cite{11,12}, among others. While the content creation capabilities of DDMs have been extensively studied, there remains a significant gap in understanding whether their generative process can simultaneously learn discriminative features. Theoretically~\cite{13}, it is expected that by tackling the challenge of precisely modeling the underlying distribution, generative models do create more comprehensive global representations for versatile downstream tasks. Hence, recently there are rising interests of training diffusion models for general-purpose applications such as classifications~\cite{63,64}. This approach, though fascinating, is extremely difficult because the complexity of diffusion models makes it challenging to capture efficient representations that benefit specific downstream tasks. 

Unlike traditional end-to-end image processing networks~\cite{b2,b6,b3}, unconditional DDMs typically possess a non-uniform architecture, with a large number of internal modules and a considerable timestep or infinitely divisible diffusion coefficients~\cite{2}. Finding an optimal layer among these internal modules and time sequences is extremely challenging, which to some extent restricts the investigation into whether the learned representations constitute a semantic latent space. In the past, some scholars have delved into this field, but they focused on the representational capacity of pre-trained models, exploring their distillable features~\cite{20,21} or zero-shot classification abilities~\cite{17,18}. Recent studies have also suggested that individual DDMs have learned manipulable bottlenecks~\cite{19}. However, none of these have addressed the core question: Can DDMs learn semantically meaningful representations?

The fundamental of our work is inspired by recent pioneer works~\cite{24} which identify the analogue between DDMs and denoising autoencoder (DAE)~\cite{25}. DAEs are a form of self-supervised learning used to extract latent structures in data, which can be understood as a reconstruction generator conditioned on corrupted images. DAEs work by eliminating several scales and small amounts of additive Gaussian noise. On the other hand, DDMs strive to eliminate multi-level noise driven by diffusion to predict possible inputs, automatically learning reconstruction features from the entire denoising Markov chain, which can be viewed as a specialized extension of multi-level DAEs that intend to model a broader underlying space distribution and generate semantically meaningful representations. From this perspective, we argue that DDMs possess powerful representations and theoretical foundations capable of understanding visual semantic content, which is not investigated before.

\begin{figure}[t]
	\centering
	\includegraphics[width=1\linewidth]{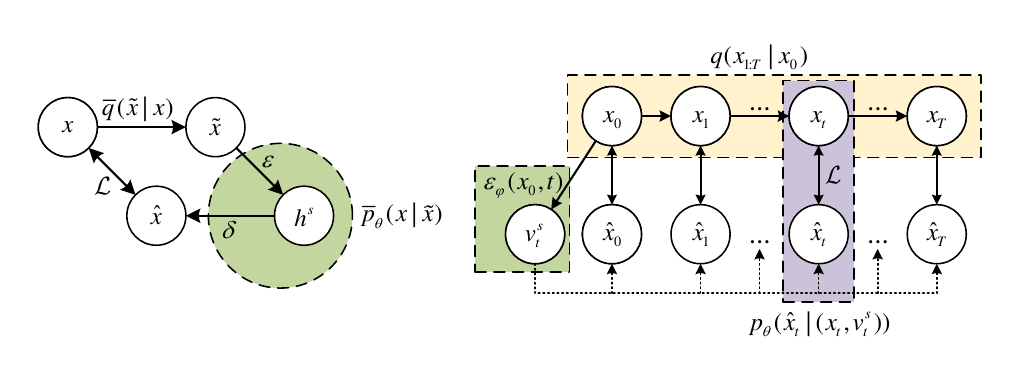}
	\vspace{-25pt}
	\caption{The analogy between a traditional DAE and our proposed DiER, multi-level DAEs derived from DDMs. (Left) A traditional DAE add (multi-scale) noise and predict the original input, the encoder $\varepsilon$ and decoder $\delta$ directly compressing corrupted images into $h^s$ to construct a latent semantic space. (Right) DiER extends DDMs by encoding a vector $v_t^s$ for each noise level, and using a single neural network with parameters $\theta$ to predict noise, thus compressing semantics of different levels into directional vectors $v_t^s$ which provide information that assists ${x_t}$ in indirectly recovering the possible original sample ${\hat{x}_t}$ from its current corrupted state, forming multi-level DAEs.}

	\label{dae_daes}
\end{figure}

To explore this area and demonstrate that DDMs can produce semantic representations, in this paper, we propose a framework for learning Diffusion Embedding Representations (DiER) at all timesteps, starting from the perspectives of self-supervision and DAEs. Instead of directly seeking one or more layers within DDMs that can generate effective representations, DiER simultaneously encodes the current timestep $t$ and the undistorted sample $x_0$ as conditional embeddings, thus transforming DDMs and the entire diffusion process into multi-level DAEs corresponding to each noise level $t$. It automatically learns image embeddings over all timesteps as self-condition. This condition consists of directional uncertainty vectors in latent compressed by the high-dimensional space, which are used for diffusion image recovery under different corruption levels. The relationship between DAE and DiER is illustrated in Fig.~\ref{dae_daes}, where $\cal L$ represents the optional loss and the distinct construction of their latent space is highlighted in the green background. Subsequently, extensive experiments are conducted on multiple benchmark datasets, comparing against the baseline and other state-of-the-art (SOTA) self-supervised methods based on generation. The goal is to validate whether the proposed method and diffusion denoising can create semantically meaningful intrinsic representations through conditional embeddings in the self-supervised generation process. The results confirm the affirmative. Moreover, we observe that the representational capabilities of this automatically learned embedding vary across different timesteps and datasets. The contributions of this paper are as follows:

\begin{itemize}
	\item We propose a feasible framework based on the diffusion model called DiER, which transforms DDMs into multi-level DAEs by encoding timesteps, enabling automatic representation learning across all timesteps.
	
	\item Extensive experiments demonstrate that the embedding representations learned by DiER can surpass SOTA self-supervised methods under small space occupancy and perform well across various datasets.
	
	\item We provide evidence that DDMs can learn discriminative representations with clear semantics. However, the quality of representations generated at different timesteps varies significantly across datasets, emphasizing the importance of accurately determining the optimal timestep when transferring DDMs to non-generative downstream tasks.
	
\end{itemize}

\section{Related Works}

\subsection{Denoising Diffusion Models for Representation Learning}

Some studies have explored whether the intrinsic representations of DDMs possess discriminative capabilities or downstream significance. However, investigations into discriminative capabilities generally do not start from scratch; instead, pre-trained large models~\cite{14,15} are utilized as zero-shot generation classifiers for generalization. They leverage vast amounts of understood information recalculating the evidence lower bound (ELBO) with reweighted values~\cite{17,18}. The entire process resembles Monte Carlo estimation. These methods do not explore whether the generated representations themselves are discriminative, so it cannot be ruled out that larger architectures and the diversity of data seen during pre-training aid the discriminative process~\cite{22,23}. Other downstream tasks presuppose that the representations of DDMs are meaningful, and although the results do confirm this assumption, they use pre-trained DDMs as teacher models, distilling internal representations into predictive decoders to bridge the gap between prediction results and groundtruth~\cite{20,21}. Similarly, studies have indicated that individual DDMs have learned manipulable latent spaces, where shifting attributes specified by the frozen bottleneck space leads to the desired changes in generated content~\cite{19}. This manipulation can also be achieved through additional self-supervised embeddings~\cite{29}.

\subsection{Generative Self-Supervision}

DAEs~\cite{25,60} are variant of autoencoders (AEs) that reconstructs original data from corrupted samples. This seminal work has inspired a series of subsequent self-supervised representation learning methods, with the most prominent being the masked autoencoder (MAE)~\cite{26}. It learns to generate pixel values of corrupted patches given visible patches and has proven to be capable of learning superior discriminative representations during the generation process. Subsequent improvements in generative self-supervised methods have been built upon this foundation, such as introducing pre-trained encoders~\cite{51} and tokenizers~\cite{53} for feature alignment or adopting local reconstruction strategies~\cite{52} and disjoint sampling~\cite{50} to enhance self-supervised efficiency. Moreover, the generation targets are no longer limited to pixels but extended to embedding vectors~\cite{28}. Compared to these high-performance or weakly supervised methods~\cite{b5}, the baseline performance of DAEs is inferior~\cite{24}. Whereas, adjusting the noise scale has been shown to improve features~\cite{61,62}. If DDMs can be transformed into multi-level DAEs, representations rich enough in semantics can be learned across such a diverse range of scales.

\section{Preliminaries}

We select the theoretical foundation of currently popular Denoising Diffusion Probabilistic Models (DDPMs)~\cite{1} as the basis for noise addition and removal algorithms. Below, we briefly introduce the background.

\subsection{Denoising Diffusion Probabilistic Models}

DDPMs, as a crucial formulation within DDMs, rely on the core concept of forward and backward stochastic Markov chains. Given a dataset $\zeta$, from which data samples $x_0$ are drawn, the forward process progressively introduces Gaussian noise, defining the distribution of its latent variables $q({x_{0:T}})$, i.e., corrupted latent variables ${x_{1:T}} = {x_1},{x_2},...,{x_T}$, where ${x_T}$ eventually degenerates to approach Gaussian noise, with $t \sim {\cal U}([0,T])$ and $T$ typically set to 1000. Each diffusion-noise sample is only correlated with the previous noise-added sample, thus the decomposition of the distribution $q$ of the noising process is as follows:
\begin{equation}
	q({x_{1:T}}|{x_0}) = \prod\limits_{t = 1}^T {q({x_t}|{x_{t - 1}})} .
\end{equation}
In this decomposition, the noise addition is parameterized by a predefined Gaussian kernel. $q({x_t}|{x_{t - 1}}) = {\cal N}({x_t};\sqrt {{\alpha _t}} {x_{t - 1}},\sqrt {1 - {\alpha _t}} \emph{\textbf{I}})$ is computed using a specified progressive timetable ${\alpha _1},{\alpha _2},...,{\alpha _T}$. The cumulative product of ${\alpha _t}$ denoted as ${\bar \alpha _t} = \prod\nolimits_{s = 1}^t {{\alpha _t}} $, formulates the marginal distribution of $q$:
\begin{equation}
	q({x_t}|{x_0}) = {\cal N}({x_t};\sqrt {{{\bar \alpha }_t}} {x_0},\sqrt {1 - {{\bar \alpha }_t}} \emph{\textbf{I}}).
\end{equation}
Thus, corrupted samples at any timestep $t$ can be derived from $x_0$. Conversely, denoising involves a backward Markov process, starting from ${\cal N}({x_T};0,\emph{\textbf{I}})$ and gradually removing the noise introduced by the diffusion process, restoring the latent variables to their original distribution:
\begin{equation}
	{p_\theta}({x_{0:T}}) = p({x_T})\prod\limits_{t = 0}^{T - 1} {{p_\theta }({x_t}|{x_{t + 1}})} ,
\end{equation}
where a neural network parameterized by $\theta$ is employed to achieve this, aiming to fully match the generative model ${p_\theta }({x_0})$ with the data distribution $q({x_0})$. To accomplish this objective, the model is trained by minimize re-weighted ELBO that fits the noise, the loss thus be written as follows:
\begin{equation}
	\label{total_loss}
	{\cal L} = {\cal L}_{recon} + {\cal L}_{prior} + {\cal L}_{simple},
\end{equation}
where these three terms are separately the reconstruction loss, the prior loss, and the simple diffusion loss~\cite{40}.

In cases where the primary learning objective does not prioritize generative capability, we are solely concerned with the model's ability to restore ${x_t}$ degraded at any timestep back to ${x_0}$. The backbone autonomously models potential distributions, and based on empirical evidence, ${\cal L}_{recon}$ and ${\cal L}_{prior}$ are omitted~\cite{3}. Denoising can be achieved by directly minimizing the discrepancy between the noise estimated by the model and the error from stochastic sampling. The optimization objective can be derived from Equation (\ref{total_loss}):
\begin{equation}
	\label{simple_loss}
	{\cal L}_{simple} = {\mathbb{E}_{x,\left\langle c \right\rangle ,\epsilon \sim {\cal N}(0,I),t}}[{w_t}\left\| { \epsilon- {\epsilon_\theta }({x_t},t,\left\langle c \right\rangle )} \right\|_2^2],
\end{equation}
where $\left\langle c \right\rangle$ is an optional conditioning, generally class labels, prompt embeddings, or other discriminative prompts~\cite{32}. ${w_t}$ denotes the weights assigned to specific timesteps, with ${w_t}$ being constant at 1 throughout this paper.

\begin{figure}[t]
	\centering
	\includegraphics[width=1\linewidth]{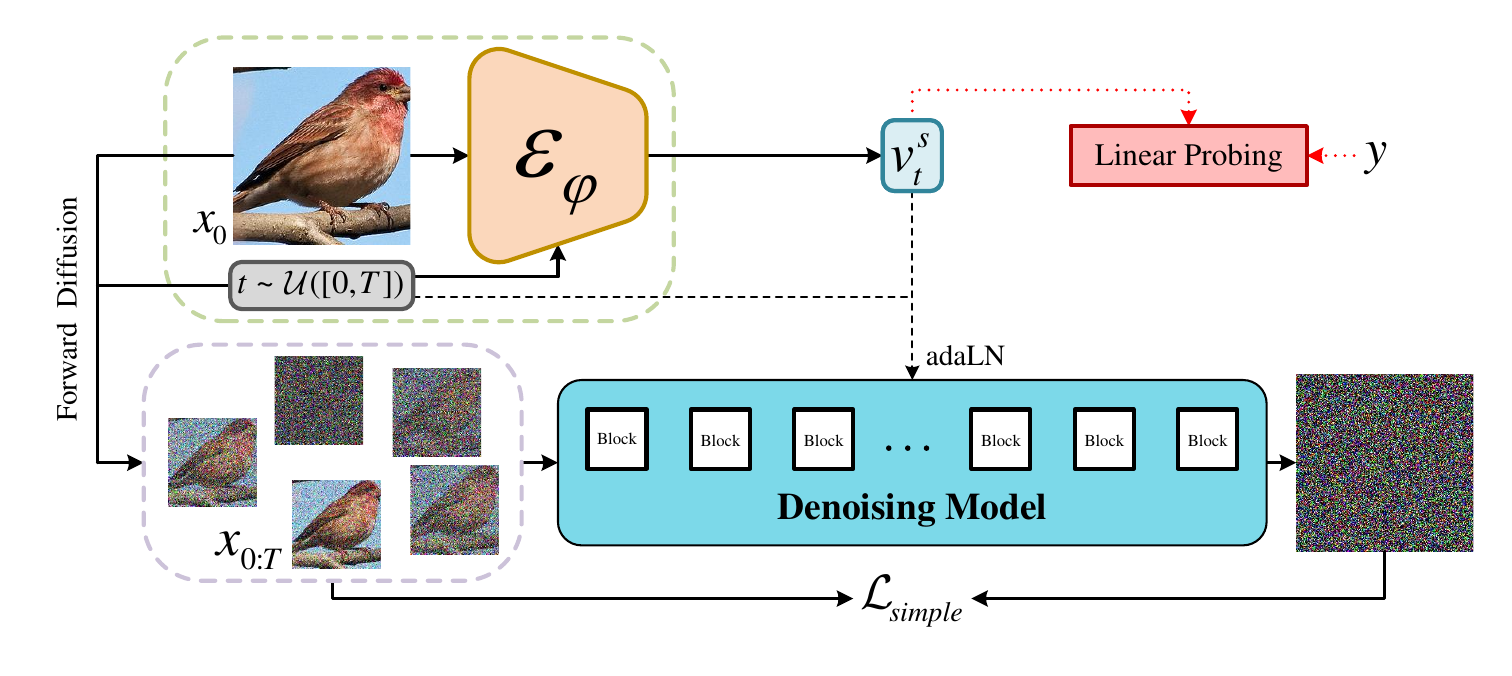}
	\vspace{-25pt}
	\caption{The pipeline used to analyze the representational capacity learning of diffusion models, where each block's internal structure remains uniform, serving for embedding through equivalent insertion computations. The encoding $v_t^s$ represents the encoding result of $x_0$ at any randomly sampled timestep $t$.}
	\label{network}
\end{figure}

\subsection{Generating Representations with Self-Supervision}

Given a dataset $\zeta$ annotated as $(x^k, y^k) \in \zeta$, where $y^k$ represents annotation or classification label of $x^k$ in the set, which is inaccessible in representation learning scheme. The goal of representation learning is to reconstruct a mapping $\varphi(x)$ that can efficiently differentiate $y^k$ in the mapped space for various inputs in a task-agnostic manner, i.e., solely based on the observation and reconstruction of $x^k$. The more consistent this representation is with the distribution of $y^k$, the less dependent we are on $y^k$~\cite{44}. Here, $\varphi(x)$ is usually implemented by a neural network. By adopting Equation (\ref{simple_loss}) above and replace the prompt conditioning $\left\langle c \right\rangle$ with the mapping that derives pseudo-annotations in place of $y^k$, we can convert this process into a self-supervision task, where the relevant features for $y^k$ can be extracted during intermediate steps amid the reconstruction of input $x^k$. In this paper, we demonstrate that this is achievable via exploring the DDMs-DAE analogy and creating a multi-level self-supervised model for efficient representation of diffusion models.

\section{Learning Semantic Embedding Representations}

Inspired by the ideas described, the overall workflow of the proposed DiER method is depicted in Fig.~\ref{network}. Specifically, we work on DDMs without external prompt $y$, and create an observation pipeline that is independent of the original denoising reconstruction process, as and employ a self-supervised model to create the representation embeddings. For each timestep $t$, a embedding representations $v_t^s$, obtained from the encoder ${\varepsilon _\varphi }$ and derived from observing $x_0$, is utilized by the diffusion model and reconstructs $x_0$ from corrupted samples $x_{1:T}$, without additional conditioning on $y$. A linear probing procedure can be then applied to analyze the representations learned by the ${\varepsilon _\varphi }$ during each denoising process, so that the quality of the representation can be measured by how good it partitions $y$ in the space.

In general, diffusion models often employ self-attention-based U-Net or other U-shaped architectures~\cite{33,34}, where coarse and fine-grained information is perceived differently across layers~\cite{36}. In this paper, we opt for Diffusion Transformers (DiT)~\cite{30} as the backbone for several reasons: 1) DiT is an architecture based on the standard Vision Transformer (ViT)~\cite{38}, which excels in capturing long-range dependencies and global contextual understanding. 2) In current open-source frameworks, DiT achieves near-optimal performance in quantitative analyses, indicating its strong generative capabilities. 3) DiT adopts a stack of homogenized modules without theoretical bottlenecks, and class embeddings do not require different scaling factors between layers, which facilitates maintaining consistency. It is noteworthy that, unlike the original version, DiT operates directly in pixel space to measure datasets of various types and resolutions.

\begin{figure}[t]
	\centering
	\includegraphics[width=0.4\linewidth]{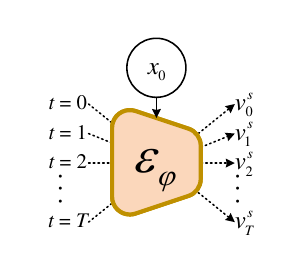}
	\vspace{-20pt}
	\caption{Based on the current $t$ in the diffusion sampling process, encode $x_0$.}
	\label{embeddings_on_timesteps}
\end{figure}

Firstly, DDMs differ from typical regression tasks in that they exhibit lower robustness and are highly sensitive to perturbations, resulting in oscillations in mean squared error~\cite{42}, making it challenging to converge to very low values. This could be attributed to the difficulty in reconstructing the diffusion strategy. Secondly, as a self-supervised generation scheme, the fundamental paradigm does not feed $c$ as supervised data, only the visible content consists of $t$, ${x_t}$, and ${x_0}$. DiER needs to learn denoising of ${x_0}$ at a single corruption level $t$ while generating semantic representations. Therefore, we regard denoising at a corruption level $t$ as a single-level DAE, assigning an embedding to each ${x_t}$, and employ a single model to learn corresponding $T$-level DAEs.

\begin{table*}[t]
	\caption{Comparison of LPA (\%) results for all methods, with the best value bold and the suboptimal value underlined}
	\label{quantitative_comparison}
	\centering
		\begin{tabular}{lcccccccc}
			\toprule
			& \multicolumn{1}{c}{\multirow{2}{*}{\textbf{MNIST}}} & \multicolumn{1}{c}{\multirow{2}{*}{\textbf{CIFAR10}}} & \multicolumn{1}{c}{\multirow{2}{*}{\textbf{BCCD}}} & \multicolumn{1}{c}{\multirow{2}{*}{\textbf{OCT2017}}} & \multicolumn{2}{c}{\textbf{CIFAR100}}             & \multicolumn{2}{c}{\textbf{Tiny-IN}}              \\
			& \multicolumn{1}{c}{}                                & \multicolumn{1}{c}{}                                  & \multicolumn{1}{c}{}                               & \multicolumn{1}{c}{}                                  & {Top-1} & Top-5                & {Top-1} & Top-5                \\ 
			\midrule
			Baseline & 62.1                                                & 51.4                                                  & 53.5                                               & 47.7                                                  & 21.8                       & 48.7                 & 10.6                       & 26.4                 \\
			
			MAE               & 94.8                                                & 50.5                                                  & 61.7                                               & 77.8                                                  & {\ul 33.1}                 & {\ul 60.8}           & {\ul 33.8}                 & {\ul 60.9}           \\
			
			CAE               & 89.3                                                & {\ul 58.9}                                            & 61.5                                               & 86.3                                                  & 27.9                       & 57.4                 & 31.1                       & 58.3                 \\
			
			DMJD              & \textbf{98.7}                                       & 58.1                                                  & \textbf{85.5}                                      & {\ul 96.9}                                            & 23.5                       & 54.4                 & 32.6                       & 59.7                 \\
			
			LMIM-ViT      & 75.1                                                & 40.7                                                  & 66.0                                               & 94.7                                                  & 15.7                       & 37.8                 & 15.9                       & 38.2                 \\
			
			LMIM-Swin     & 87.8                                                & 49.7                                                  & 58.9                                               & 79.9                                                  & 28.8                       & 55.9                 & 16.8                       & 39.3                 \\
			
			MAGE              & {\ul 96.5}                                          & 54.8                                                  & 38.2                                               & 85.9                                                  & 30.0                       & 57.7                 & 20.8                       & 44.2                 \\
			
			DiER              & \makecell{89.9$_{400}$}                                         & \textbf{59.7$_{100}$}                                  & {\ul 67.9$_{400}$}                                  & \textbf{98.1$_{500}$}                                  & \textbf{37.7$_{100}$}       & \textbf{67.3$_{100}$} & \textbf{36.7$_{100}$}       & \textbf{62.5$_{100}$}\\
			\bottomrule
		\end{tabular}
\end{table*}

Specifically, we simultaneously optimize an encoder ${\varepsilon _\varphi }$, which encodes the initial sample ${x_0}$ into a semantic direction-agnostic vector space~\cite{29}. In practice, the semantic capacity of a single vector is insufficient, akin to learning a universal embedding for $T$ DAEs. As illustrated in Fig.~\ref{embeddings_on_timesteps}, the simultaneous encoding of $x_0$ and the current denoising timestep $t$ is performed. Thus, we ultimately learn a set of vector embeddings $\{ v_0^s,v_1^s,...v_T^s\} \subset {\mathbb{R}^d}$, dependent on the entire diffusion process, to guide denoising for all ${x_t}$, where $d$ is the dimensionality, determined by the DiT submodule embedding configuration. At any $t$, the purpose of $v_t^s = {\varepsilon _\varphi }({x_0},t)$ is to extract intrinsic auxiliary information corresponding to timestep $t$ from ${x_0}$, automatically learned by DiER to compress high-dimensional data into latent vectors with uncertainty. This information varies in direction for different data types, possibly representing fine-grained details or rough overall sample structures, and is embedded into self-attention layers and multi-layer perceptrons through adaptive layer norm (adaLN)~\cite{43}. The final optimization objective is rewritten from Equation (\ref{simple_loss}) as:
\begin{equation}
	\small
	\label{final_loss}
	{\cal L}_{simple} = {\mathbb{E}_{x,{\varepsilon _\varphi }({x_0},t) ,\epsilon \sim {\cal N}(0,I),t}}[{w_t}\left\| { \epsilon- {\epsilon_\theta }({x_t},t,{\varepsilon _\varphi }({x_0},t) )} \right\|_2^2].
\end{equation}
We allow DiT and ${\varepsilon _\varphi }$ to browse unlabeled data $\zeta$ and attempt reconstruction using embeddings learned by ${\varepsilon _\varphi }$. Once DiT accurately reconstructs the additive noise itself, $v_t^s$ serves as the self-conditional embedding representation of diffusion across $T$ timesteps.

By adopting Equation (\ref{final_loss}), we enforce the encoder ${\varepsilon _\varphi }$ to learn a low-dimensional mapping to reflect the regularity of the differentiation information between $x_0$ and $x_t$, i.e., the denoising rules, for each timestep $t$, over the entire dataset. It can be naturally expected that summarizing the denoising rules for the entire dataset approaches semantically meaningful knowledge about how specific images are generated, e.g., how to draw images belonging to some classes or possessing some features. Simultaneously, it is equivalent to the ${\cal L}$ in multi-level DAEs depicted in Fig.~\ref{dae_daes}, as it attempts to indirectly reconstruct the original image from the corrupted samples by estimating the intermediate noise added. In this way, we achieve the transformation from DDMs to $T$-level DAEs and a series of potentially meaningful embedded representations for the DDMs. Next, we will further validate the semantic expressive power of $v_t^s$.

\begin{figure*}[t]
	\centering
	\includegraphics[width=1\linewidth]{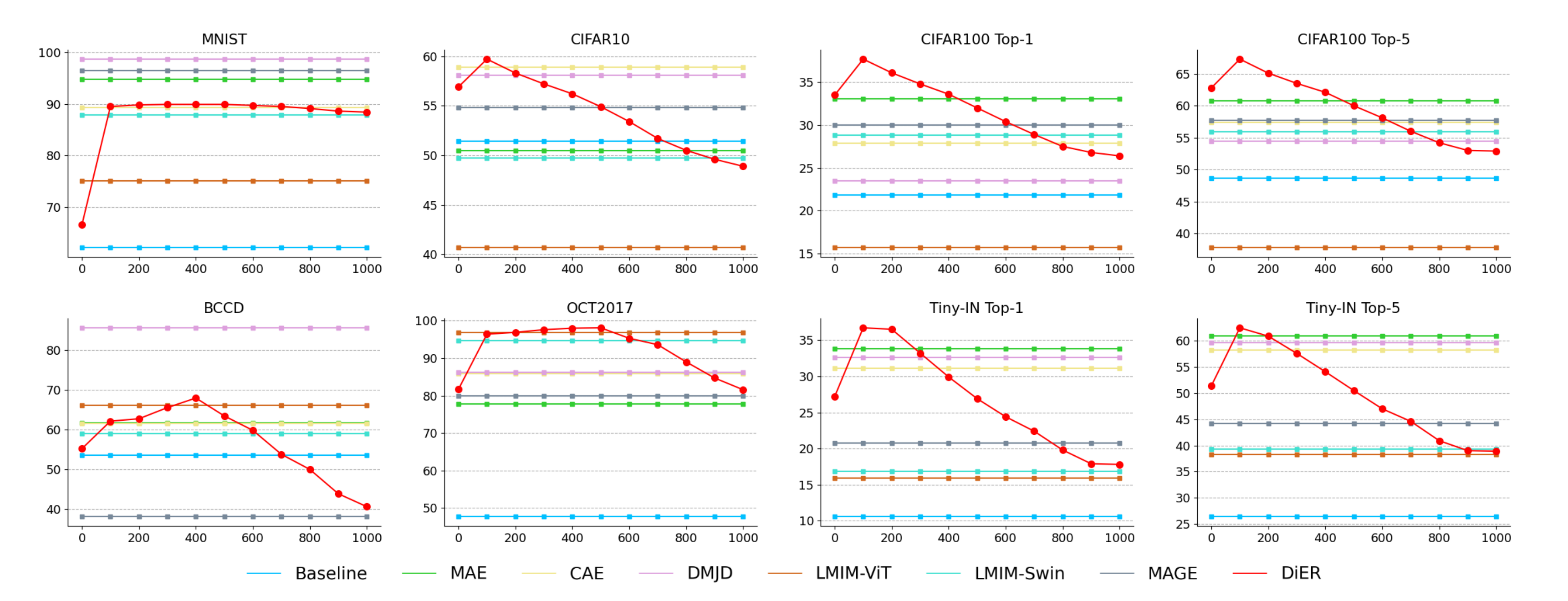}
	\caption{The LPA results of DiER across all timesteps are depicted. The x-axis represents the timestep, while the y-axis denotes the LPA. The representations were evaluated every 100 timesteps, with the maximum $t$ being 999. Solid lines indicate the Top-1 Acc., while dashed lines represent the Top-5 Acc.}
	\label{timestepcmp}
\end{figure*}

\begin{table*}[t]
	\caption{The test results of DiER across all LPA (\%) settings and rounded for neatness. The bolded values represent the optimal accuracy achieved at $t$. For MNIST, it is distinguished by the second decimal place.}
	\label{dier_timestep}
	\centering
		\begin{tabular}{l|cccccccccccc}
			\toprule
			Datasets                                        & $t$   & 0    & 100           & 200  & 300  & 400           & 500           & 600  & 700  & 800  & 900  & 999  \\ \cline{2-13} 
			\textbf{MNIST}                                          &       & 66.5 & 89.5          & 89.8 & 89.9 & \textbf{89.9} & 89.9          & 89.7 & 89.5 & 89.1 & 88.6 & 88.4 \\
			\textbf{CIFAR10}                                        &       & 56.9 & \textbf{59.7} & 58.3 & 57.2 & 56.2          & 54.9          & 53.4 & 51.7 & 50.5 & 49.6 & 48.9 \\
			\textbf{BCCD}                                           &       & 55.2 & 62.1          & 62.7 & 65.5 & \textbf{67.9} & 63.4          & 59.8 & 53.8 & 50.0 & 43.9 & 40.7 \\
			\textbf{OCT2017}                                        &       & 81.7 & 96.4          & 96.9 & 97.6 & 98.0          & \textbf{98.1} & 95.3 & 93.6 & 89.0 & 84.7 & 81.6 \\
			\multicolumn{1}{c|}{\multirow{2}{*}{\textbf{CIFAR100}}} & Top-1 & 33.5 & \textbf{37.7} & 36.1 & 34.8 & 33.6          & 32.0          & 30.4 & 28.9 & 27.5 & 26.8 & 26.4 \\
			\multicolumn{1}{c|}{}                          & Top-5 & 62.8 & \textbf{67.3} & 65.1 & 63.5 & 62.1          & 60.0          & 58.1 & 56.0 & 54.2 & 53.0 & 52.9 \\
			\multicolumn{1}{c|}{\multirow{2}{*}{\textbf{Tiny-IN}}}  & Top-1 & 27.2 & \textbf{36.7} & 36.5 & 33.2 & 29.9          & 26.9          & 24.4 & 22.4 & 19.8 & 17.9 & 17.8 \\
			\multicolumn{1}{c|}{}                          & Top-5 & 51.4 & \textbf{62.5} & 60.9 & 57.6 & 54.1          & 50.5          & 47.0 & 44.6 & 40.9 & 39.0 & 38.9\\
			\bottomrule
		\end{tabular}
\end{table*}

\section{Experiments}

\subsection{Implementation Details}
In this section, extensive experiments were conducted to evaluate the proposed method by comparing it with the baseline and other SOTA self-supervised representation learning methods. We will first introduce the datasets used, the comparison methods, and the baseline settings.

\paragraph{Datasets}
To validate the effectiveness of representation learning across various scales of ${y^k}$, we selected multiple datasets with diverse content and different resolutions. The datasets involved include MNIST~\cite{46}, CIFAR10~\cite{45}, CIFAR100~\cite{45}, Tiny-IN~\cite{47}, BCCD\footnote{\url{https://github.com/Shenggan/BCCD_Dataset}}, and OCT2017~\cite{48}. For specific details, please refer to Appendix~\ref{subsectionDataset}.

\paragraph{Methods for Comparison}
We will compare the quality of embeddings learned by the proposed DiER with baseline and established self-supervised representation learning methods. The baseline is defined as Diff-AE~\cite{29}, as it initially employs vanilla DDMs and uses the downsampling part of the diffusion U-Net as the encoder for the semantic learner, resulting in a manipulable linear latent. The representation of the encoded vectors obtained from this space will be analyzed in the experiments. For self-supervised representation learning, we chose the pioneering MAE~\cite{26}, as well as other SOTA works, including CAE~\cite{51}, DMJD~\cite{50}, LMIM~\cite{52}, and MAGE~\cite{53}. LMIM has shown improvements on both Swin Transformer~\cite{54} and ViT~\cite{38}, so we will test the effectiveness of both approaches.

\paragraph{Experiment Details} 
For each method, our preference is to resize images smaller than $32 \times 32$ to $32 \times 32$, leave images with a resolution of $64 \times 64$ unchanged, and uniformly resize larger images to either $224 \times 224$ or $256 \times 256$. If it is necessary to use a pre-trained preprocessing model (such as CAE using the DALL-E tokenizer~\cite{55}) or to extract multi-level patches (such as DMJD), images are resized to sizes compatible with their encoding capacity, rather than the predefined sizes aforementioned above. To ensure fairness and exclude interference from data augmentation, all methods only use random horizontal flipping, with color jittering and zooming cropping disabled. For more detailed settings and architecture hyperparameters of DiER and the baseline on each dataset, please refer to Appendix~\ref{network_specifications}.

\subsection{Experimental Results and Analysis}

\paragraph{Linear Probe Accuracy}

We quantitatively validate the performance of each method using the commonly employed metric for self-supervised learning, linear probe accuracy (LPA)~\cite{26}. LPA measures the semantic quality of learned representations, with higher values indicating better performance. For datasets with 10 or fewer classes, we evaluate only the Top-1 LPA; for datasets with more classes, we test both Top-1 and Top-5 accuracies. Detailed results are presented in Table~\ref{quantitative_comparison}, where subscripts indicate the timestep at which the highest LPA is achieved.

It can be observed that on the four datasets, DiER consistently SOTA self-supervised methods at the optimal timestep. Specifically, on CIFAR10, it outperforms CAE by 0.8\%, on OCT2017 it outperforms DMJD by 1.2\%, and on CIFAR100 it leads by a significant margin, with Top-1 and Top-5 LPA surpassing the closest competitor MAE by 4.6\% and 6.5\%, respectively. On Tiny-IN, both Top-1 and Top-5 surpass MAE by 2.9\% and 1.6\%. However, performance on other datasets is not optimal; for instance, on BCCD, it lags behind DMJD by approximately 17.6\%, and on MNIST, it falls short of MAGE by 6.6\%. Nevertheless, the proposed method demonstrates consistent performance across all datasets, regardless of the number of classes, yielding acceptable results rather than excelling on only one dataset. This indicates that DDMs can acquire semantic embedding representations through the automatic learning of conditional embeddings, and the advantages in generation can extend to subsequent tasks, rendering them scalable visual models.

\begin{figure*}[t]
	\centering
	\includegraphics[width=1\linewidth]{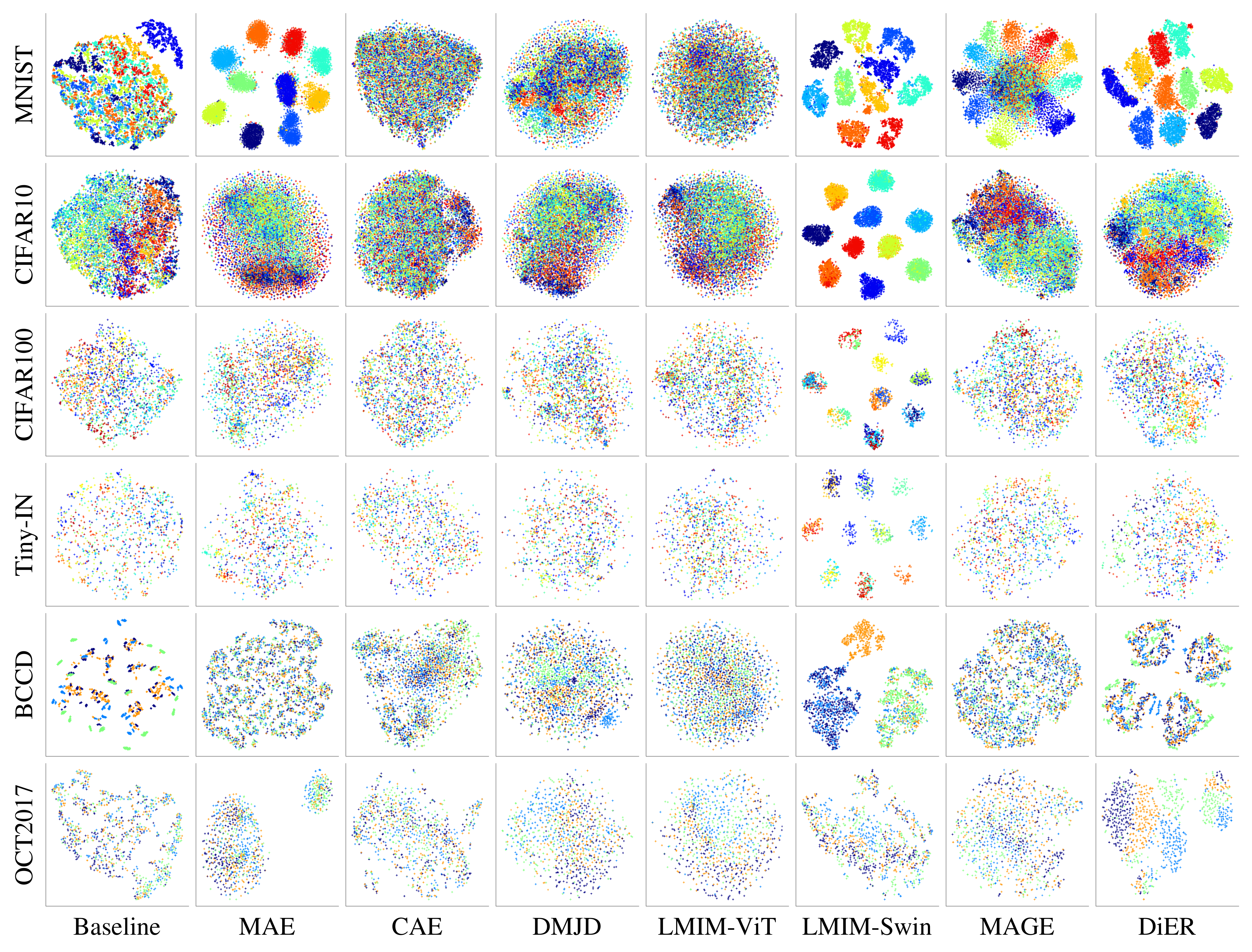}
	\caption{The t-SNE visualization of representations extracted from test images of all datasets, where each class is represented by a distinct color.}
	\label{tsnecmp}
\end{figure*}

\paragraph{Timestep-Dependent Coding}

Firstly, DiER significantly outperforms the baseline, leading by more than 15\% in all cases except for CIFAR10 and BCCD. This improvement may stem from the proposed method encoding auxiliary embeddings for each level DAE at finer timesteps, rather than a single multi-scale embedding. Moreover, DiT eliminates the need for multiple scaled embedding vectors without concern for information loss during integral downsampling and upsampling, thereby enhancing representational capacity. This result also highlights a phenomenon where the optimal representations mostly occur at intermediate timesteps ($v_{100}^s$, $v_{400}^s$, $v_{500}^s$), rather than at $v_{0}^s$ or $v_{T}^s$. We will present the LPA of our method at all tested timesteps later accompanied by further discussion.

\paragraph{Comparative Analysis Across All Timesteps}
\label{Comparative_All_Timesteps}

The comparative analysis of DiER across all timesteps is illustrated in Fig.~\ref{timestepcmp}, with specific numerical values detailed in Table~\ref{dier_timestep}. Generally, it tends to achieve optimal performance within specific timestep ranges, centered around a certain peak value of $t$, with decreasing trends observed on both sides. Although the peak performance of the LPA typically surpasses that of most SOTA methods, determining the optimal $t$ swiftly remains a challenge.

We conjecture that optimal timesteps phenomenon arises because excessively low timesteps would cause DDMs to degrade into conventional AEs, where the reconstruction of corrupted pixels becomes nearly trivial, thereby yielding less informative embeddings. Conversely, higher timesteps lead to severe corruption levels in the samples, resulting in the destruction of finer details. To enhance the restorative capabilities of the latent space, individual embeddings may need to accommodate the reconstruction of images under various degraded states, thereby encoding more granular, context-driven information. This observation aligns with established self-supervised learning theories, wherein both excessively high and low image corruption rates lead to a decline in representation learning effectiveness~\cite{24,26}. Additionally, we hypothesize that the optimal $t$ for representation learning is related to the continuity of individual sample content. For CIFAR10, CIFAR100, and Tiny-IN datasets, where a significant portion of the pixels is relevant to the classification task (i.e., the proportion occupied by the classification target to the overall image), the optimal representations occur at $v_{100}^s$. Conversely, for the remaining datasets, the optimal representations are observed at $v_{400}^s$, $v_{500}^s$. Among these, MNIST predominantly features a black background, BCCD involves cell classification with only a small portion of the pixels occupied by cells need to be classified, and OCT2017 exhibits similar characteristics in the lesion regions.

\paragraph{Compressible Representation}

Similar to other self-supervised methods, the embedding representations learned in the proposed vanilla encoder exhibit compressibility (other self-supervised methods typically use \texttt{cls token} or \texttt{avgpooling} to obtain vectors for classification). The default adaLN embedding vector length of DiER's backbone is 1024, indicating that even when compressing the preceding features into a very small shape for classification, it still achieves outstanding LPA. Thus, the DiER encoder framework is scalable. If the encoder architecture is modified and combined with other approaches, it can be utilized to learn latent representations in higher dimensions and more complex manifolds.

\paragraph{Visual Analysis}

Utilizing t-SNE~\cite{58} to visualize the normalized output of high-dimensional features encoded by all methods, as shown in Fig.~\ref{tsnecmp} for comparison. Due to the large number of classes in CIFAR100 and Tiny-IN, only samples from 20 randomly selected classes are used for visualization. It is worth noting that the separation produced by models does not necessarily indicate precise semantic representation, as they may erroneously mix samples from different classes (e.g., LMIM-Swin). Conversely, a mixed t-SNE does not necessarily imply poor performance, it may also result from the inability of low-dimensional manifolds to accurately project data in high-dimensional space. However, models that preserve correct aggregation results in low dimensions undoubtedly possess good semantic representation capabilities. Across all experiments, DiER maintains a strong clustering trend, even on datasets with 100 or 200 classes. The embedding representations of DiER across all tested timesteps are visualized and can be found in Appendix~\ref{Visualization_of_Embedding}. The recovery outcomes from the random Gaussian noise and stochastic code, as well as the discussion of relationship between generative capability and representations of the model, are elaborated in Appendix~\ref{Recovering_Samples}.

\section{Discussion and Conclusion}

Efficient representation learning for diffusion models not only energizes their surging potential as generic foundation models, but also help with developing interpretable and controllable architectures that enhance the quality of data generation. At the core of our DiER lies the modeling of diffusion timesteps interfering in the encoding space, automatically predicting the diffusion embedding conditions at the same timestep without any additional data augmentation. By resolving the backbone of the diffusion process as a series of DAEs, our work explores and harnesses the tremendous potential of DDMs as a self-supervised model, reaching an efficient timestep-dependent series of representation for this type of model. Through pre-training on six datasets, extensive LPA testing, and comparison with current SOTA self-supervised approaches, the results demonstrate the effectiveness of DiER’s design. We also analyze the possible reasons for the optimal representations of different content datasets at different timesteps, visualize the embedding representation manifold space, and discuss the compression and scalability of current encoder-generated representations.

\paragraph{Limitation}

We discuss potential limitations of this work. Initially, the objective of this work is not to create new SOTA self-supervised learning schemes but rather to explore the potential of DDMs as representation learning models for better utilization. Moreover, the design of DiT with an appended vanilla encoder is quite memory-intensive and computationally expensive. Most importantly, finding an optimal representation in such a long diffusion process is challenging, uniformly sampling timesteps at a large interval is advisable, yet downstream tasks should not waste so much time. Further work should focus on efficient $t$ selection, novel encoder architectures, ways to couple the encoder and backbone, and implementation of downstream tasks, which are our next priorities.

\section*{Acknowledgements}
	This study was funded by the National Natural Science Foundation of China (Grant No. U22A2041).

{
	\small

	\bibliographystyle{unsrt}
	
	\bibliography{cites.bib}
	\medskip
}

\newpage
\appendix

In Appendix~\ref{ExperimentSetting}, we provide detailed information about the experiments, including datasets, hyperparameters, and network architectures. Appendix~\ref{Visualization_of_Embedding} offers visualization and analysis of these embedding representations. In Appendix~\ref{Recovering_Samples}, we demonstrate DiER's ability to recover from various types of corrupted samples and discuss the correlation between generation capability and semantic representation.

\section{Experiment Setting}
\label{ExperimentSetting}

\subsection{Dataset Elaborations}
\label{subsectionDataset}

\begin{table}[h]
	\centering
	\caption{The specifications of the datasets used in this paper.}
		\label{Dataset_Specification}
				\resizebox{0.5\textwidth}{!}{
	\begin{tabular}{lcccccc}
		\toprule
		Dataset & MNIST & CIFAR10 & CIFAR100 & Tiny-IN & BCCD & OCT2017 \\
		\midrule
		Resolution & $28 \times 28$ & $32 \times 32$ & $32 \times 32$ & $64 \times 64$ & $640 \times 480$ & $256 \times 256$ \\
		Class count & 10 & 10 & 100 & 200 & 4 & 4 \\
		Type & Numeral & General & General & General & Medical & Medical \\
		\bottomrule
	\end{tabular}
}
\end{table}

Table~\ref{Dataset_Specification} provides a brief description of the datasets used. MNIST is a dataset for handwritten digit recognition, consisting of 10 classes with images of resolution $28 \times 28$. CIFAR10 and CIFAR100 are commonly used benchmark classification datasets with 10 and 100 classes, respectively, and images of resolution $32 \times 32$. Tiny-IN is a subset of the ImageNet dataset~\cite{49}, comprising 200 classes with images of resolution $64 \times 64$. BCCD and OCT2017 are both simple and fundamental medical imaging datasets, each with 4 classes. They have relatively high resolutions, $640 \times 480$ and $256 \times 256$ respectively.

\begin{table}[t]
	
	\caption{Architecture of the backbone and encoder utilized in DiER. Here, Patch denotes the patch size defined by the transformer, Hidden indicates the dimension of the hidden layers, Heads represents the number of heads in the multi-head self-attention mechanism, Base denotes the base channel number per layer in the encoder, Blocks signifies the number of residual blocks at the same resolution, Attention refers to the resolution queues employing self-attention, and Channel* denotes the base channel amplification factor per layer for vanilla encoders or U-Net.}
	\label{DiER_structure}
	\centering
	\resizebox{0.5\textwidth}{!}{
		\begin{tabular}{llcccccc}
			\toprule
			\multicolumn{1}{l}{}                           &           & \textbf{MNIST}          & \textbf{CIFAR10}        & \textbf{CIFAR100}       & \textbf{Tiny-IN}        & \textbf{BCCD}             & \textbf{OCT2017}          \\ \hline
			\multicolumn{1}{c}{\multirow{6}{*}{Backbone}} & Input     & $32 \times 32$ & $32 \times 32$ & $32 \times 32$ & $64 \times 64$ & $256 \times 256$ & $256 \times 256$ \\
			\multicolumn{1}{c}{}                          & Patch     & $2 \times 2$   & $2 \times 2$   & $2 \times 2$   & $4 \times 4$   & $8 \times 8$     & $8 \times 8$     \\
			\multicolumn{1}{c}{}                          & Hidden    & 192            & 192            & 192            & 384            & 768              & 768              \\
			\multicolumn{1}{c}{}                          & Depth     & 12             & 12             & 12             & 12             & 16               & 16               \\
			\multicolumn{1}{c}{}                          & Heads     & 3              & 3              & 3              & 6              & 12               & 12               \\
			\multicolumn{1}{c}{}                          & MLP Ratio & 4              & 4              & 4              & 4              & 4                & 4                \\ 
			\midrule
			\multicolumn{1}{c}{\multirow{5}{*}{Encoder}}  & Base      & 128            & 128            & 128            & 128            & 256              & 256              \\
			\multicolumn{1}{c}{}                          & Blocks    & 2              & 2              & 2              & 2              & 2                & 2                \\
			\multicolumn{1}{c}{}                          & Attention & (16,8)         & (16,8)         & (16,8)         & (16,8)         & (16,8)           & (16,8)           \\
			\multicolumn{1}{c}{}                          & Heads     & 4              & 4              & 4              & 4              & 4                & 4                \\
			\multicolumn{1}{c}{}                          & Channel*  & (1,2,3,4)      & (1,2,3,4)      & (1,2,3,4)      & (1,2,3,4)      & (1,1,2,2,3,4)    & (1,1,2,2,3,4)    \\ 
			\bottomrule
		\end{tabular}
	}
\end{table}

\begin{table}[t]
	\caption{The architecture employed in the baseline utilizes a configuration where the vanilla encoder shares a similar setup with the downsampling portion of the U-Net backbone.	}
	\label{Diffae_structure}
	\centering
	\resizebox{0.5\textwidth}{!}{
	\begin{tabular}{c|ccccccc}
		\toprule
		\multirow{2}{*}{}& \multirow{2}{*}{Input}            & \multirow{2}{*}{Base} & \multirow{2}{*}{Blocks} & \multirow{2}{*}{Attention} & \multirow{2}{*}{Head} & Backbone  & \multirow{2}{*}{Channel*} \\ \cline{7-7}
		&                                   &                       &                         &                            &                       & Encoder   &                           \\
		\midrule
		\multirow{2}{*}{\textbf{MNIST}}    & \multirow{2}{*}{$32 \times 32$}   & \multirow{2}{*}{128}  & \multirow{2}{*}{2}      & \multirow{2}{*}{(16,8)}    & \multirow{2}{*}{4}    & \multicolumn{2}{c}{(1,1,2,3,4)}                                                                         \\
		\cline{7-8}
		&                                   &                       &                         &                            &                       & \multicolumn{2}{c}{(1,1,2,3,4,4)}                                                                       \\
		\midrule
		\multirow{2}{*}{\textbf{CIFAR10}}  & \multirow{2}{*}{$32 \times 32$}   & \multirow{2}{*}{128}  & \multirow{2}{*}{2}      & \multirow{2}{*}{(16,8)}    & \multirow{2}{*}{4}    & \multicolumn{2}{c}{(1,1,2,3,4)}                                                                         \\
		\cline{7-8}
		&                                   &                       &                         &                            &                       & \multicolumn{2}{c}{(1,1,2,3,4,4)}                                                                       \\
		\midrule
		\multirow{2}{*}{\textbf{CIFAR100}} & \multirow{2}{*}{$32 \times 32$}   & \multirow{2}{*}{128}  & \multirow{2}{*}{2}      & \multirow{2}{*}{(16,8)}    & \multirow{2}{*}{4}    & \multicolumn{2}{c}{(1,1,2,3,4)}                                                                         \\
		\cline{7-8}
		&                                   &                       &                         &                            &                       & \multicolumn{2}{c}{(1,1,2,3,4,4)}                                                                       \\
		\midrule
		\multirow{2}{*}{\textbf{Tiny-IN}}  & \multirow{2}{*}{$64 \times 64$}   & \multirow{2}{*}{128}  & \multirow{2}{*}{2}      & \multirow{2}{*}{(16,8)}    & \multirow{2}{*}{4}    & \multicolumn{2}{c}{(1,1,2,3,4)}                                                                         \\
		\cline{7-8}
		&                                   &                       &                         &                            &                       & \multicolumn{2}{c}{(1,1,2,3,4,4)}                                                                       \\
		\midrule
		\multirow{2}{*}{\textbf{BCCD}}     & \multirow{2}{*}{$256 \times 256$} & \multirow{2}{*}{128}  & \multirow{2}{*}{2}      & \multirow{2}{*}{(32)}      & \multirow{2}{*}{4}    & \multicolumn{2}{c}{(1,1,2,2,3,3,4,4)}                                                                   \\
		\cline{7-8}
		&                                   &                       &                         &                            &                       & \multicolumn{2}{c}{(1,1,2,2,3,3,4,4)}                                                                   \\
		\midrule
		\multirow{2}{*}{\textbf{OCT2017}}  & \multirow{2}{*}{$256 \times 256$} & \multirow{2}{*}{128}  & \multirow{2}{*}{2}      & \multirow{2}{*}{(32)}      & \multirow{2}{*}{4}    & \multicolumn{2}{c}{(1,1,2,2,3,3,4,4)}                                                                   \\
		\cline{7-8}
		&                                   &                       &                         &                            &                       & \multicolumn{2}{c}{(1,1,2,2,3,3,4,4)}  \\
		\bottomrule                                                                
	\end{tabular}
}
\end{table}

\subsection{Network Specifications}
\label{network_specifications}

For all self-supervised methods, regardless of the input image resolution, experiments on MNIST, CIFAR10, and CIFAR100 datasets are conducted using configurations of relevant architectures with tiny blocks (such as ViT-T and Swin-T). Tiny-IN adopts a small setup, while BCCD and OCT2017 utilize a base setup. This adjustment is also applied to Diff-AE and DiER, where changes in the settings and depth of the DiT submodule and encoder are made during experimentation. All experiments are conducted on a server equipped with 6 NVIDIA RTX A6000 GPUs. Pre-training is performed using a single GPU with a fixed random seed. Each dataset undergoes 200 epochs of training from scratch, and testing is conducted using the established settings of each method in its paper.

Frameworks for DiER and the baseline on different datasets are meticulously designed, as illustrated in Table~\ref{DiER_structure} and Table~\ref{Diffae_structure}. Similarly, for MNIST, CIFAR10, and CIFAR100, DiT-T is employed as the backbone, while DiT-S is used for Tiny-IN, and DiT-B is utilized for BCCD and OCT2017. To validate the effectiveness, we deepen the layers of the baseline as much as possible. Both DiER and the baseline employ vanilla encoders, where the learned features from the final layers are transformed into embedding vectors using adaptive average pooling for insertion into the backbone. Meanwhile, the vector length is set to 1024 and 512 for DiT and Diff-AE by default, respectively. DiER uniformly scales the vectors to the specified hidden dimensions, whereas the baseline scales them based on the channel multiplication coefficients of different residual layers.

\begin{figure}
	\centering
	\includegraphics[width=0.5\textwidth]{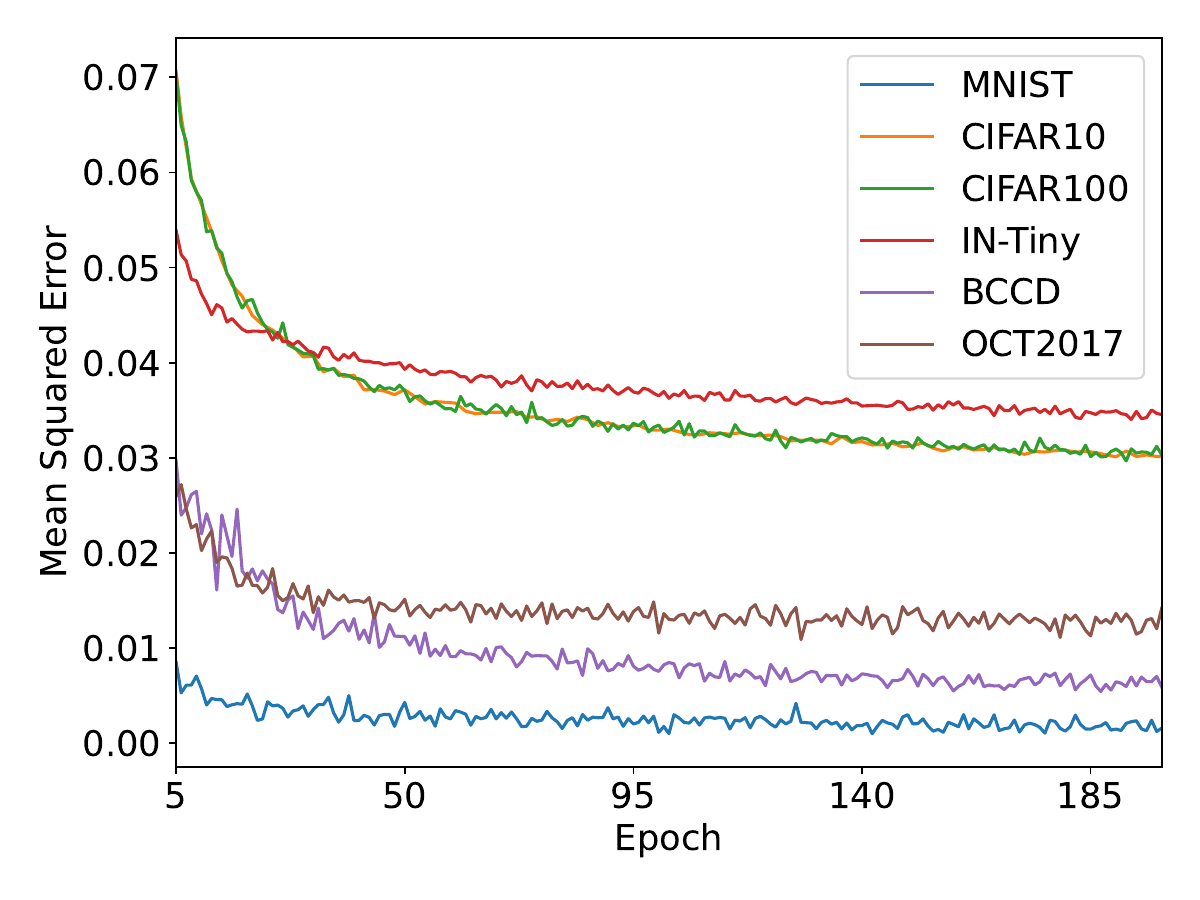}
	\caption{The trend of mean squared error reduction during the training process over all datasets.}
	\label{diffusion_pretrain_loss}
\end{figure}

During the diffusion process, both DiER and the baseline follow the same schedule, consisting of 1000 timesteps using the linear strategy. Pre-training lasts for 200 epochs and employs the Adam optimizer with default hyperparameters~\cite{56} and a constant learning rate. Higher learning rates lead to gradient collapse; hence, a learning rate of $1 \times 10^{-5}$ is set for BCCD and OCT2017, while for other datasets, it is set to $1 \times 10^{-4}$. The pre-training process as illustrated in Fig.~\ref{diffusion_pretrain_loss}. LPA testing spans 100 epochs, utilizing the AdamW optimizer with a weight decay of 0.05~\cite{57}. The maximum learning rate is twice that of pre-training, with a minimum learning rate of 0, a warm-up period of 10 epochs, and a linear decay strategy. Furthermore, since $T=1000$, LPA testing is conducted 11 times at intervals of 100 timesteps from 0 to 999 to evaluate the current LPA of DiER at timestep $t$. This ensures uniform testing of time-dependent embedding representations throughout the diffusion process. All results are recorded, and the highest score along with the current timestep is considered the final accuracy.

\begin{figure*}[t]
	\centering
	\includegraphics[width=1\linewidth]{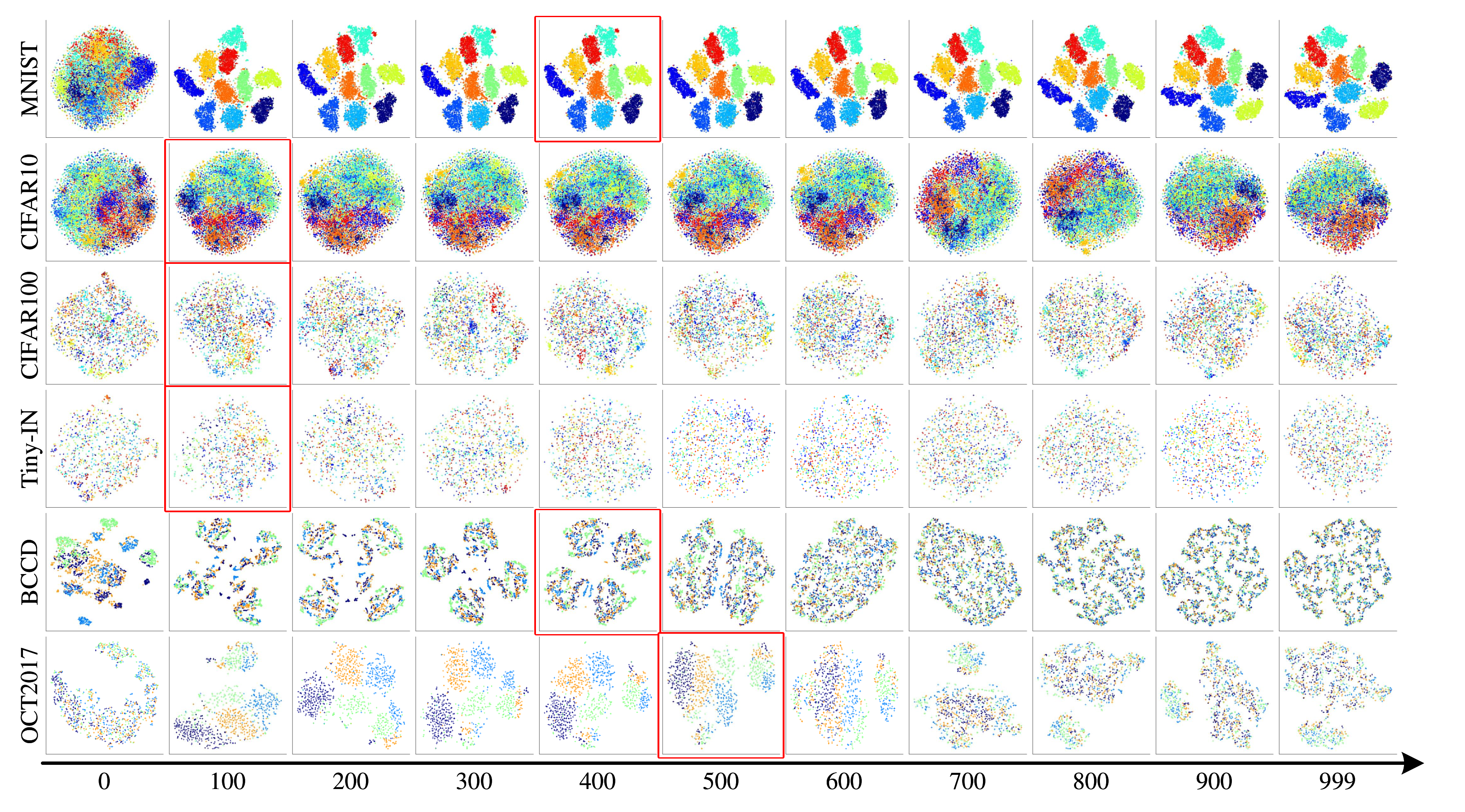}
	\caption{Embedding representations are visualized at intervals of 100 timesteps, with the optimal LPA highlighted within a red box.}
	\label{dierrepresentationvisual}
\end{figure*}

\section{Visualization of Embedding Representations with Small Margins}
\label{Visualization_of_Embedding}

Similarly, at intervals of 100 timesteps, all embeddings of the test set samples were collected and subjected to t-SNE dimensionality reduction analysis, as depicted in Fig.~\ref{dierrepresentationvisual}. For certain types of datasets, such as MNIST and CIFAR10, the features aggregate into distinct clusters with minimal changes over time. However, datasets like BCCD and OCT2017 exhibit drastic changes in clustering patterns with increasing timesteps. This observation aligns well with the trends observed in our previous LPA evaluations, where the accuracy levels are positively correlated with the degree of clustering in visualizations. This correlation is particularly pronounced in the case of OCT2017, where accuracy inevitably declines after reaching optimal representations. Furthermore, except for MNIST, the general state of $v_0^s$ remains relatively favorable, which may be attributed to the self-supervised representation capabilities of noise-free vanilla AEs.

\section{Recovering Samples from Random Noise and Stochastic Code}
\label{Recovering_Samples}

\begin{figure*}[p]
	\centering
	
	\includegraphics[width=1\linewidth]{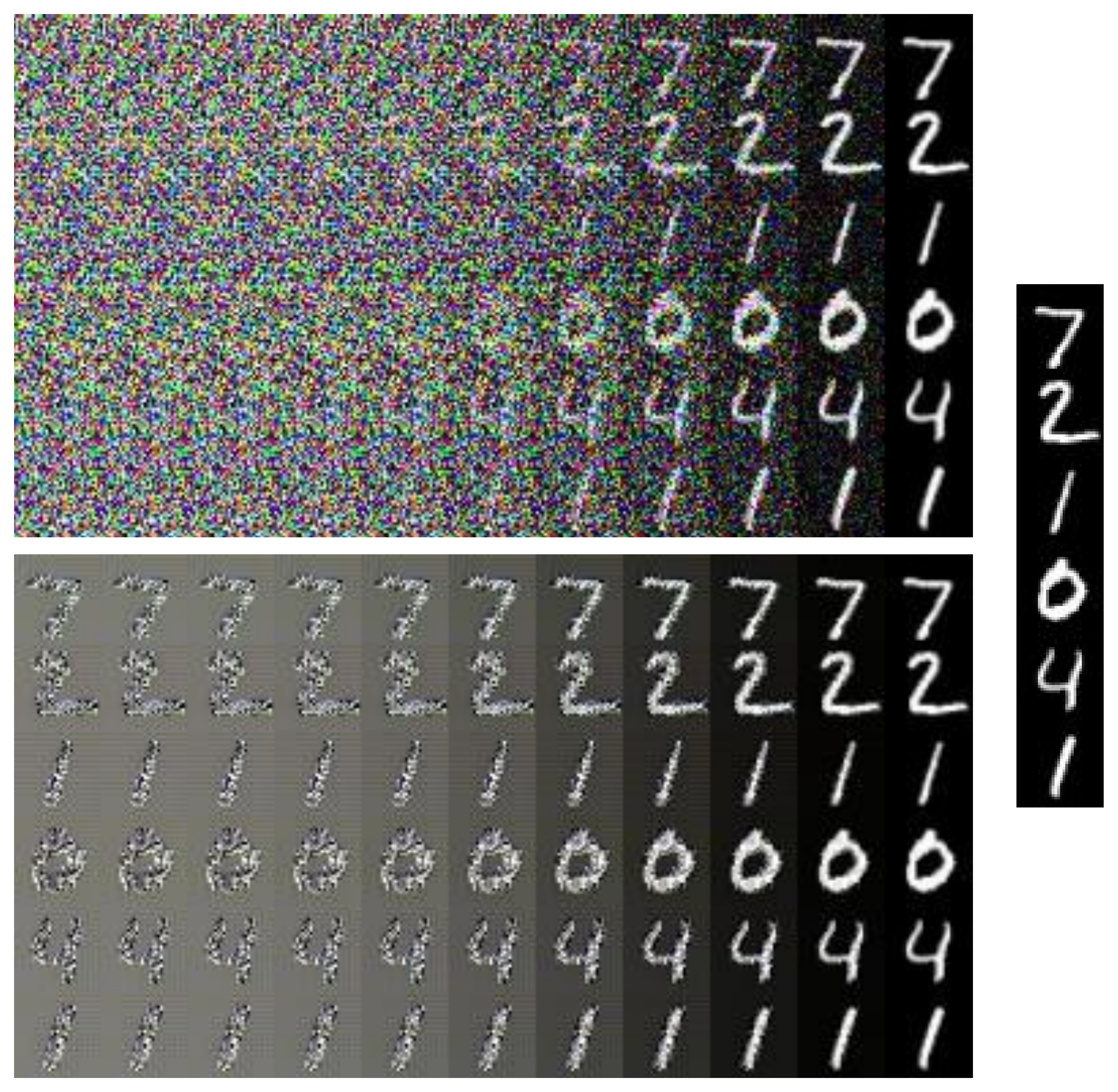}
	\caption{The results of recovering MNIST test samples from random noise and stochastic codes.}
	\label{mnistrecoverfromnoise}
\end{figure*}

\begin{figure*}[p]
	\centering
	\includegraphics[width=1\linewidth]{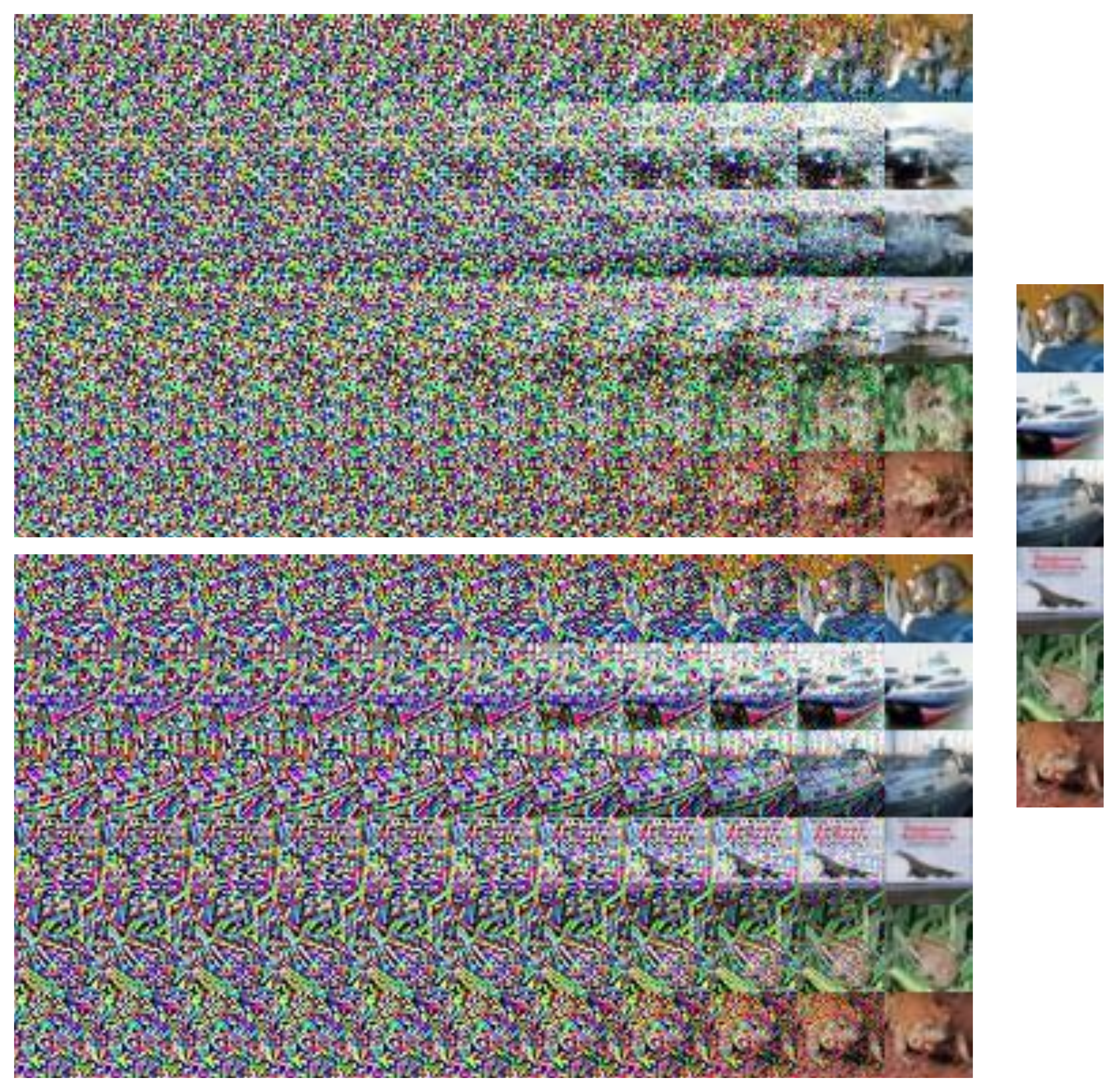}
	\caption{The results of recovering CIFAR10 test samples from random noise and stochastic codes.}
	\label{cifar10recoverfromnoise}
\end{figure*}

\begin{figure*}[p]
	\centering
	\includegraphics[width=1\linewidth]{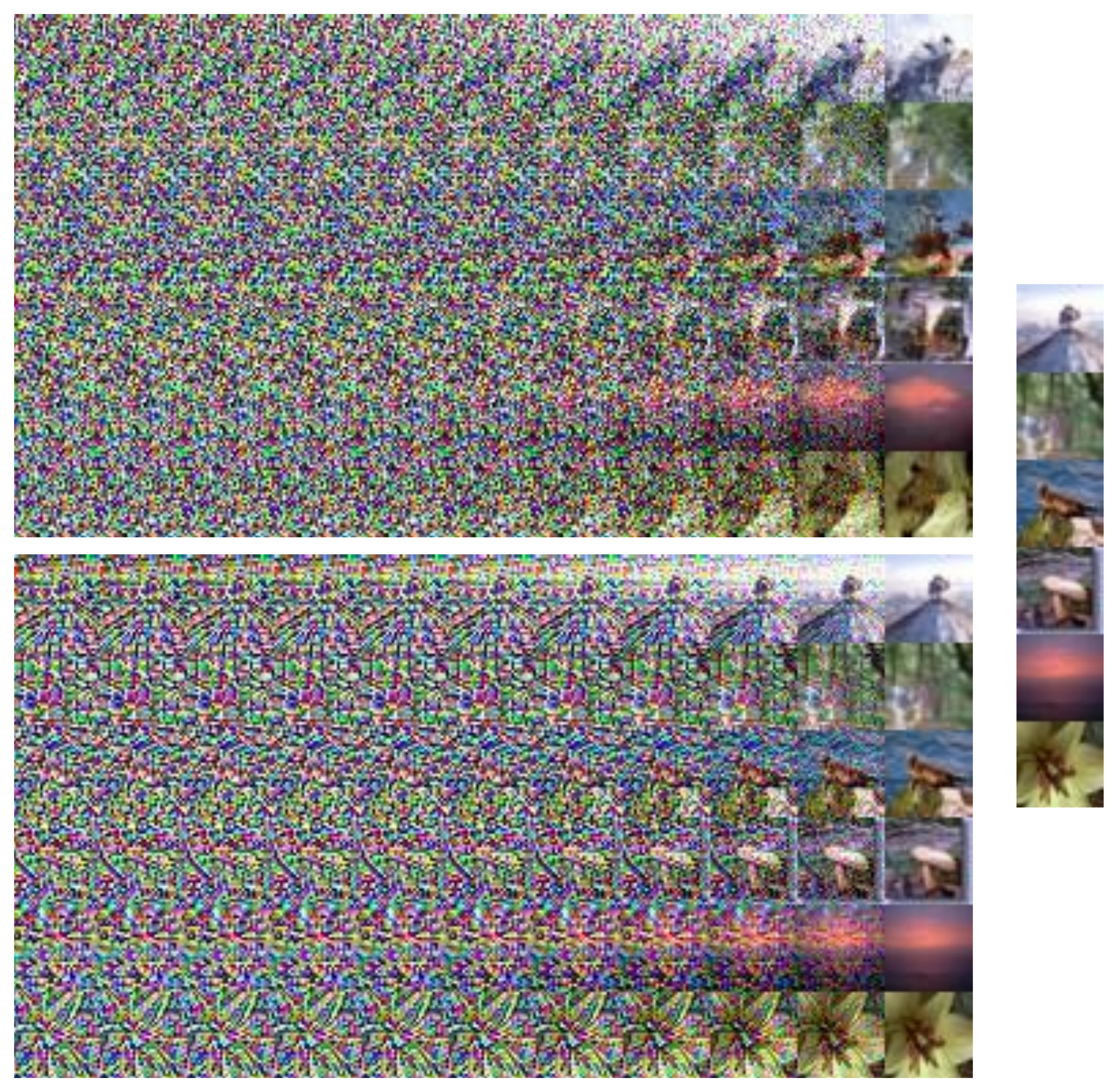}
	\caption{The results of recovering CIFAR100 test samples from random noise and stochastic codes.}
	\label{cifar100recoverfromnoise}
\end{figure*}

\begin{figure*}[p]
	\centering
	\includegraphics[width=1\linewidth]{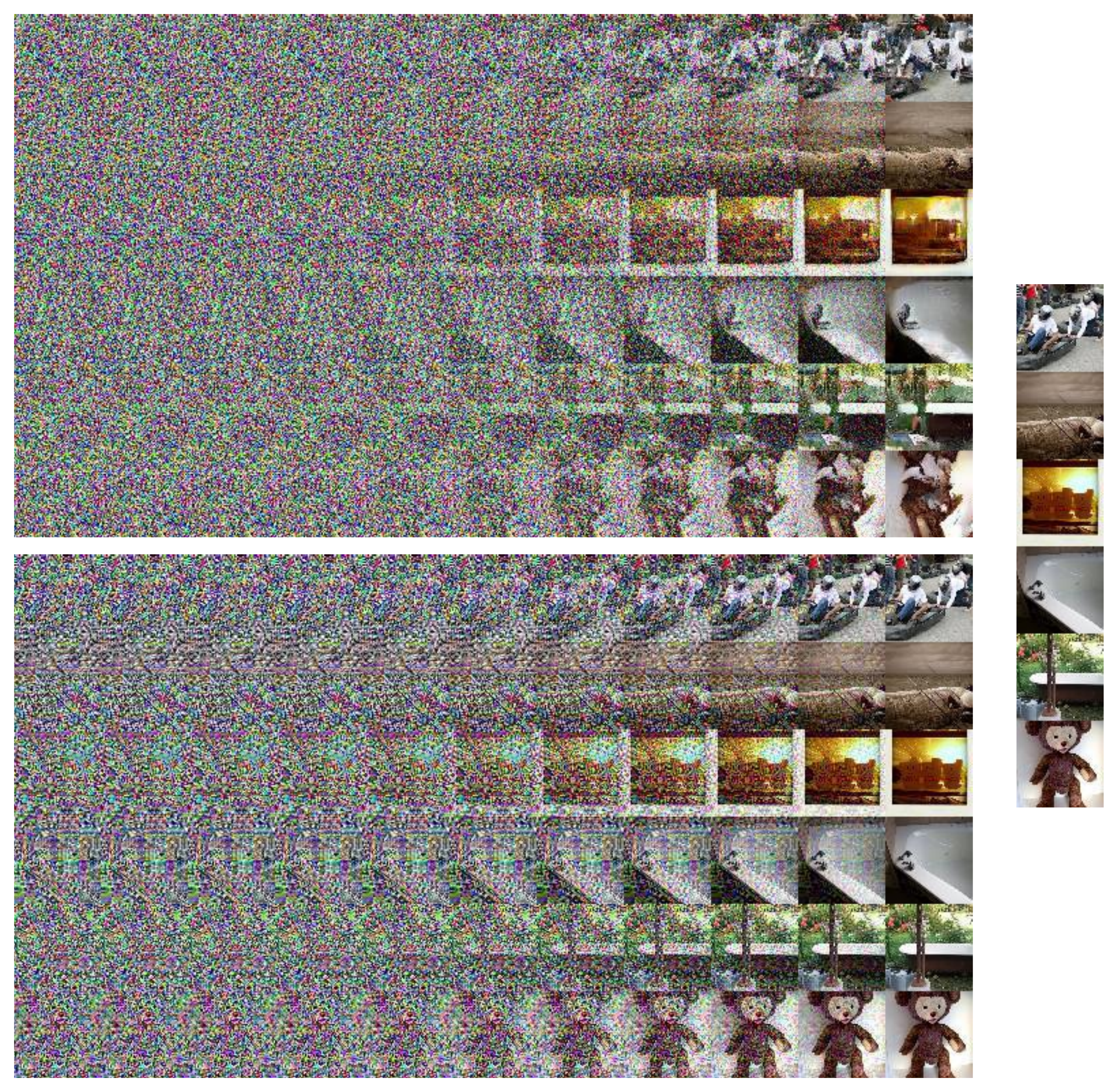}
	\caption{The results of recovering Tiny-IN test samples from random noise and stochastic codes.}
	\label{tinyinrecoverfromnoise}
\end{figure*}

\begin{figure*}[p]
	\centering
	\includegraphics[width=1\linewidth]{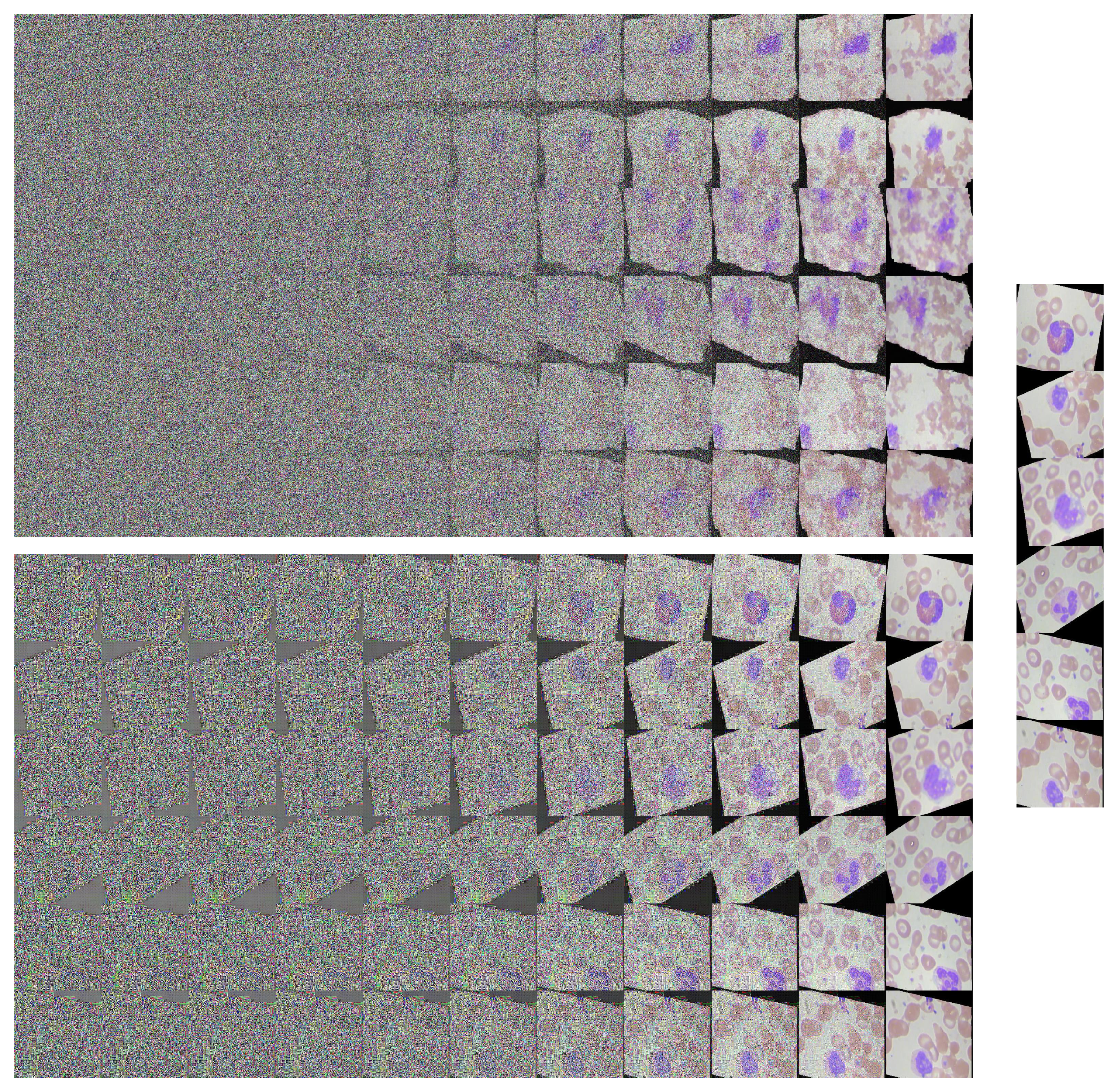}
	\caption{The results of recovering BCCD test samples from random noise and stochastic codes.}
	\label{bccdrecoverfromnoise}
\end{figure*}

\begin{figure*}[p]
	\centering
	\includegraphics[width=1\linewidth]{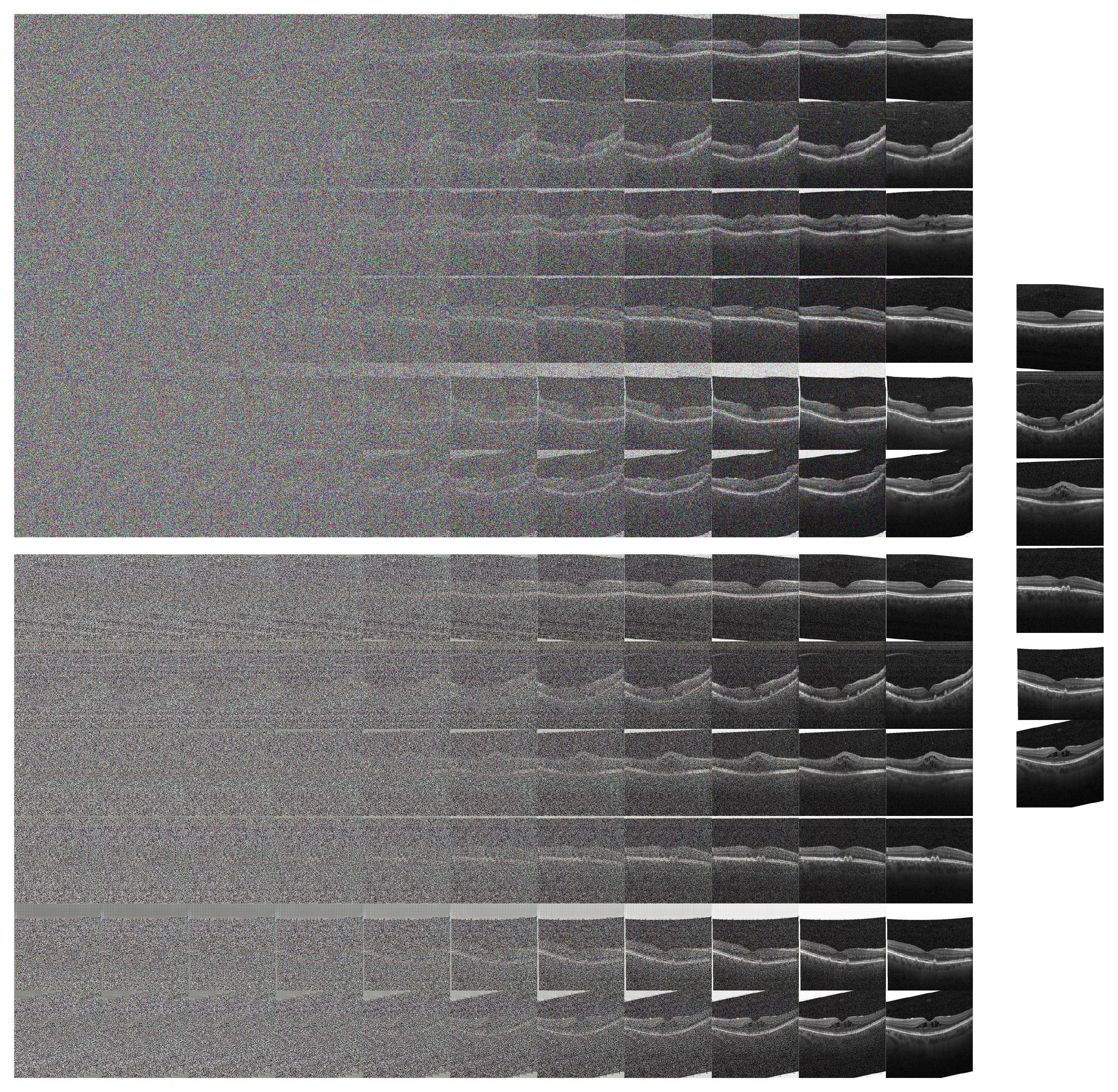}
	\caption{The results of recovering OCT2017 test samples from random noise and stochastic codes.}
	\label{oct2017recoverfromnoise}
\end{figure*}

As a qualitative evaluation, we conducted a study on the recovery of original images from random noise and stochastic codes. The random noise was generated using random Gaussian sampling, similar to the diffusion process, while stochastic coding treated the diffusion process as a form of encoding, with noise derived from the model-calculated noisy encoding~\cite{29}. The comparisons across six datasets are illustrated in Figs.~\ref{mnistrecoverfromnoise},~\ref{cifar10recoverfromnoise},~\ref{cifar100recoverfromnoise},~\ref{tinyinrecoverfromnoise},~\ref{bccdrecoverfromnoise},~\ref{oct2017recoverfromnoise}. Samples recovered from random noise are positioned in the top left of each figure, those recovered from stochastic codes are in the bottom left, and groundtruth images are displayed on the right. It is evident that DiER can accurately reconstruct images from stochastic codes, but the limited information contained in the 1024-length vector makes it challenging to embed more details, resulting in poorer outcomes for samples involving random noise.

Generally, the ability to reconstruct images from partial information signifies the generative model's capacity for scalable visual representation learning~\cite{44}. For instance, in CIFAR100 and Tiny-IN, DiER struggles to estimate the original image categories from random noise, only capturing rough outlines. However, this observation is not absolute, while the recovery performance in MNIST is the best, its LPA lags behind that of the some comparison methods. This discrepancy is related to the inherent information content of the images. For example, in the case of OCT2017, although DiER fails to perfectly reconstruct groundtruth from random noise, the basic pathological classifications of the recovered images show minimal deviation, indicating robust representation capabilities.

\clearpage

\clearpage

\clearpage

\clearpage

\clearpage

\clearpage

\clearpage

\end{document}